\documentclass[letterpaper, 10 pt, conference]{ieeeconf}  

\IEEEoverridecommandlockouts                              

\usepackage{amsmath} 
\usepackage{amssymb}  
\usepackage[usenames, dvipsnames, table]{xcolor}
\usepackage[noadjust]{cite}
\usepackage{breqn}
\usepackage{overpic}

\usepackage{caption}
\usepackage{subcaption}

\usepackage{algorithm}     
\usepackage{algorithmic}   

\usepackage{tikz}
\usepackage{textcomp}
\usepackage{hyperref}
\usepackage{lipsum}

\newcommand\copyrighttext{%
  \footnotesize \textcopyright 2021 IEEE. Personal use of this material is permitted.
  Permission from IEEE must be obtained for all other uses, in any current or future
  media, including reprinting/republishing this material for advertising or promotional
  purposes, creating new collective works, for resale or redistribution to servers or
  lists, or reuse of any copyrighted component of this work in other works.}
\newcommand\copyrightnotice{%
\begin{tikzpicture}[remember picture,overlay]
\node[anchor=south,yshift=10pt] at (current page.south) {\fbox{\parbox{\dimexpr\textwidth-\fboxsep-\fboxrule\relax}{\copyrighttext}}};
\end{tikzpicture}%
}

\usepackage[markup=bfit, deletedmarkup=sout, authormarkup=superscript]{changes}
\definechangesauthor[name={ar}, color=teal]{AR}
\definechangesauthor[name={NN}, color=MidnightBlue]{NN}
\definechangesauthor[name={ce}, color=ForestGreen]{CE}
\definechangesauthor[name={ms}, color=red]{MS}
\definechangesauthor[name={mw}, color=purple]{MW}
\definechangesauthor[name={jk}, color=blue]{JK}

\def\SPSB#1#2{\rlap{\textsuperscript{{#1}}}\SB{#2}}

\def\SB#1{\textsubscript{{#1}}}

\makeatletter
\let\NAT@parse\undefined
\makeatother

\usepackage{hyperref}
\usepackage[capitalize]{cleveref}

\hypersetup{
  linkcolor  = BrickRed,
  citecolor  = Green,
  urlcolor   = NavyBlue,
  colorlinks = true,
}

\hypersetup{
  linkcolor  = black,
  citecolor  = black,
  urlcolor   = black,
  colorlinks = true,
}

\title{\LARGE \bf Contact Anticipation for Physical Human--Robot Interaction with Robotic Manipulators using Onboard Proximity Sensors}

\author{Caleb Escobedo$^{*}$,  Matthew Strong, Mary West, Ander Aramburu, Alessandro Roncone
\thanks{All authors are with the Department of Computer Science, University of Colorado Boulder, 1111 Engineering Drive, Boulder, CO USA {\tt\small name.surname@colorado.edu}}
\thanks{* Corresponding author.}
}

\begin{document}

\maketitle
\copyrightnotice

\thispagestyle{empty}
\pagestyle{empty}

\begin{abstract}
In this paper, we present a framework that unites obstacle avoidance and deliberate physical interaction for robotic manipulators.
As humans and robots begin to coexist in work and household environments, pure collision avoidance is insufficient, as human--robot contact is inevitable and, in some situations, desired.
Our work enables manipulators to anticipate, detect, and act on contact. 
To achieve this, we allow limited deviation from the robot's original trajectory through velocity reduction and motion restrictions. Then, if contact occurs, a robot can detect it and maneuver based on a novel dynamic contact thresholding algorithm. 
The core contribution of this work is dynamic contact thresholding, which allows a manipulator with onboard proximity sensors to track nearby objects and reduce contact forces in anticipation of a collision. Our framework elicits natural behavior during physical human--robot interaction. We evaluate our system on a variety of scenarios using the Franka Emika Panda robot arm; collectively, our results demonstrate that our contribution is not only able to avoid and react on contact, but also \textsl{anticipate} it. 
\end{abstract}


\section{Introduction}\label{sec:intro}

Robots have begun to transition from assembly lines, where they are separated from humans, to environments where human–robot interaction is inevitable \cite{villani2018survey}. With this shift, research in physical human–robot interaction (pHRI) has grown  to allow robots to work with and around humans on complex tasks. Safe pHRI requires robots to both avoid harmful collisions and continue to work toward their main task, whenever possible. Furthermore, robots must reliably sense their surrounding environment and parse pertinent information in real-time to avoid potentially harmful collisions.

\begin{figure}
    \setlength{\fboxrule}{0.25mm}%
    \makeatletter
    \@wholewidth0.25mm
    \makeatother
    \vspace{+8pt}
    \begin{overpic}[width=0.90\linewidth,percent]{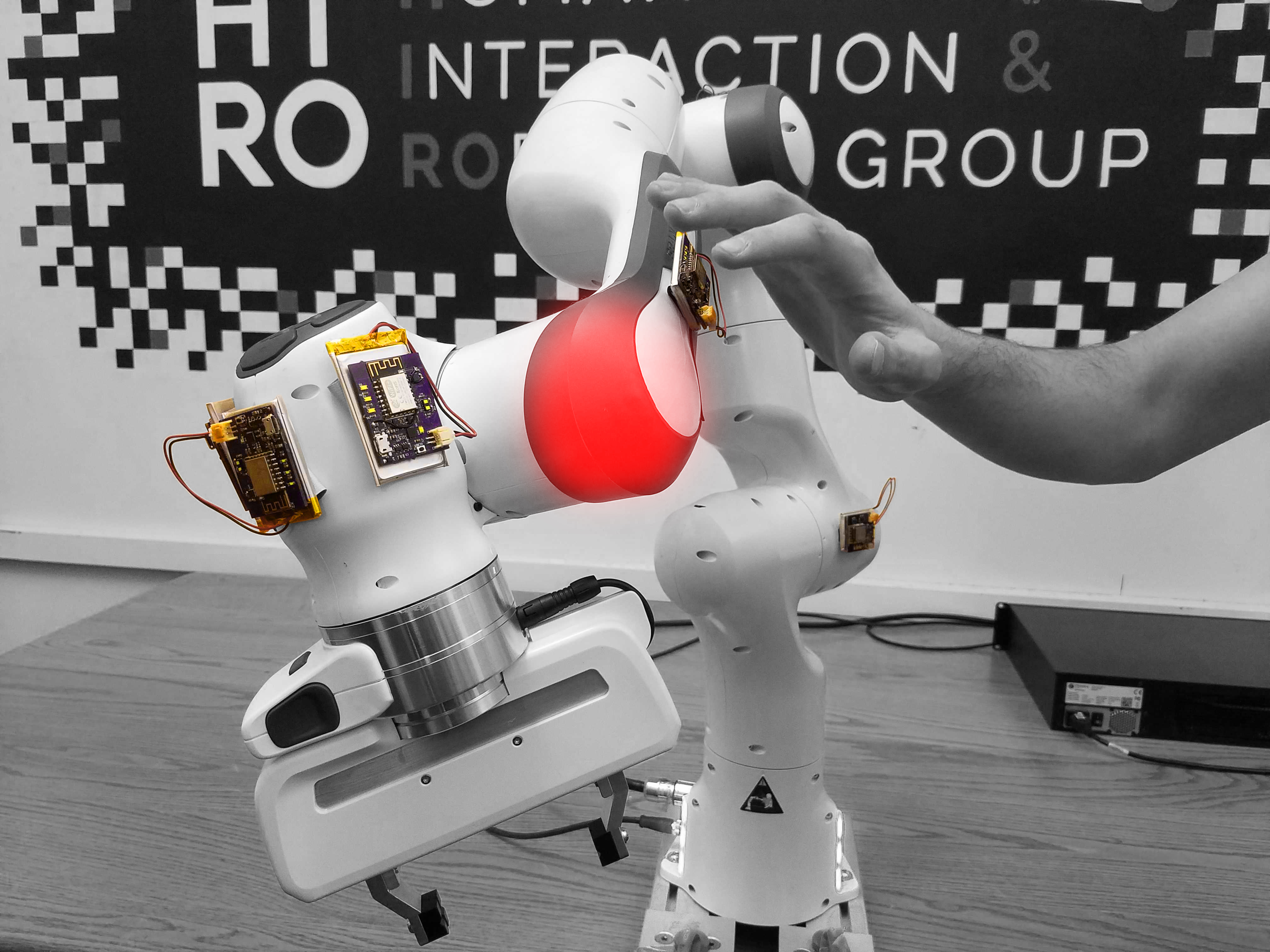}%
        \put(74,-5){\frame{\includegraphics[width=0.32\linewidth]{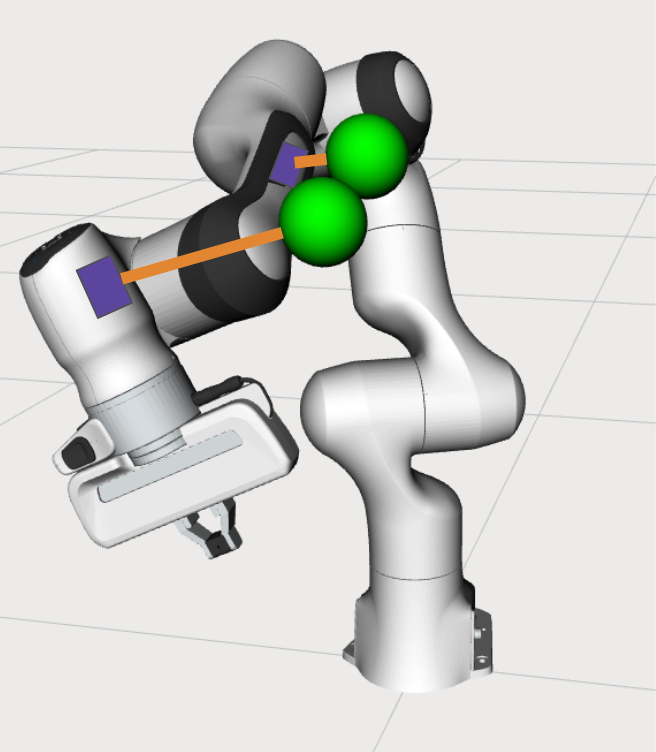}}}%
    \end{overpic}\vspace{+7pt}
    \caption{Four sensor units (shown in color) detecting a human hand near the robot's surface. The bottom right image shows a visualization of sensed objects (green spheres) based on values from proximity sensors. The red portion of the robot shows where contact is expected to be made (an interpolation of the data from the proximity sensors).}
    \label{fig:SU_Contact}\vspace{-16pt}
\end{figure}
However, as HRI scenarios become commonplace, resorting to pure collision avoidance is not sufficient---contact is inevitable, and, at times, desirable. For example, when humans work in close proximity, if contact needs to be made, a single nudge or tap can cause a person to move away and create more space for movement. We take a similar approach in this work: when contact is made, the robot moves slightly away to allow free movement near the contact area. Previous work (e.g. \cite{villani2018survey, lasota2017survey}) outline research done in both collision avoidance and contact detection independently with no work focusing on the transition between the two.

The first step to \textsl{anticipating potential contact} is to accurately perceive objects in a robot's nearby space. One common solution consists of depth sensing cameras, which are externally mounted and can perceive both humans and objects in order to allow a robot to avoid unwanted collisions.
These systems are computationally expensive, prone to occlusion, and low-bandwidth---they typically operate at a maximum of $30$ Hz and are thus unsuitable for highly dynamic environments and the presence of humans. While \cite{melchiorre2019collison} proposed solving the occlusion problem with multiple cameras, their approach struggles to provide real-time avoidance.
When a robot interacts with the environment, there exists an inverse relationship between the frequency of occlusions and proximity to the interaction---that is, the closer a human and robot operate, the more frequently occlusions occur.
In all, these limitations make it challenging for a robot to guarantee safety in unstructured, dynamic environments.
To mitigate this issue, in this work we utilize custom-built artificial skin prototypes equipped with proximity sensors to anticipate collisions and increase sensitivity to contact. As detailed in \cref{fig:SU_Contact}, multiple sensors are distributed along the robot's body, so as to enable a robot to perceive its surroundings in real-time.
With whole-body sensing, the robot can observe otherwise visually occluded objects in its nearby space.

We posit that environmental information identified by onboard proximity sensors enables new robot behaviors that go beyond pure avoidance. With this in mind, we introduce a framework outlined in \cref{fig:framework} that allows a robot manipulator to anticipate, detect, and act on contact: 1) when an object approaches, we use whole-body sensing to track its position with respect to the whole body of the robot; 2) potentially harmful collisions are first avoided (by means of the kinematic redundancy of the robot) and then anticipated (in order to reduce contact force); 3) when contact happens, small external forces are recognized by the robot and used to alter its trajectory in real time.
To control the robot we formulate our main task as a quadratic programming (QP) optimization problem. With this formulation the robot avoids collision with reduced velocities and relaxed avoidance constraints to allow for intentional contact. Furthermore, we devise a novel contact detection and reaction scheme to register anomalies in noisy external force data and leverage onboard sensors to increase sensitivity to contact.
We experimentally evaluate the proposed framework on a real robot in three different scenarios. Altogether, our results demonstrate that onboard sensing and dynamic contact thresholding can allow for a smooth transition between avoidance and contact. By leveraging these contributions, we are one step closer to allowing humans and robots to safely operate in close proximity.

This paper is organized as follows: \cref{sec:related_work} provides background on environmental sensing and safety in HRI through obstacle avoidance methods.
\cref{sec:materials_and_methods} details the entire system setup for this framework.
Specifically, \cref{sec:dynamic_contact_thresholds} introduces the main contribution, dynamic contact thresholding for contact anticipation and identification.
In \cref{sec:experiment_design}, we outline our system evaluation with static and dynamic environments, followed by results and discussion of our experimental scenarios in \cref{sec:results}. A video demonstration of this work is available at \href{https://youtu.be/hg4LLBKKV6I}{https://youtu.be/hg4LLBKKV6I}.
\section{Related Work}\label{sec:related_work}
Work in safe HRI has largely focused on \textsl{complete} obstacle avoidance, leading to zero contact between manipulators and humans.
For example, \cite{vogel2013projection, Svarny_2019} demonstrates the use of designated safety zones around a robot where the robot slows down and eventually stops its movement entirely to avoid a collision. 
Similarly, dynamic avoidance methods utilize a externally mounted 3D camera to inform the robot's movement in dynamic environments \cite{tulbureclosing, nascimentocollision, flacco2012depth}. 
\cite{nguyen2018compact} showcases avoidance with on-board vision and 3D human skeleton estimation and visualizes potential contact in the robot's peripersonal space \cite{roncone2015learning}.
Approaches in \cite{zanchettin2015safety, lasota2014toward} embrace the idea of working in close proximity; however, these methods rely on external sensing and maintain a safe distance away from the human at all times, not permitting contact. 
Some recent works have introduced on-board perception with proximity sensors (see e.g. \cite{ding2019proximity,mittendorfer2015realizing,watanabe2021calibration}). However, the proximity data is typically used to solely inform avoidance  \cite{avanzini2014safety,ding2020collision}, and not the transition to contact.

On the other hand, work on collision detection and reaction has generally not integrated prior-to-contact sensing. Several recent works focused on collision detection; none of these implementations proactively perceived the environment to anticipate contact \cite{geravand2013human, mariotti2019admittance, magrini2016hybrid, magrini2014estimation, haddadin2017robot, haddadin2008collision}. The framework presented in \cite{de2012integrated} is the most similar work ours, in that it allows for both avoidance and contact---although it lacks a smooth transition between the two.

 \section{Method}\label{sec:materials_and_methods}
\begin{figure*}\vspace{+5pt}
  \centering
    \includegraphics[width=\linewidth, height=7cm]{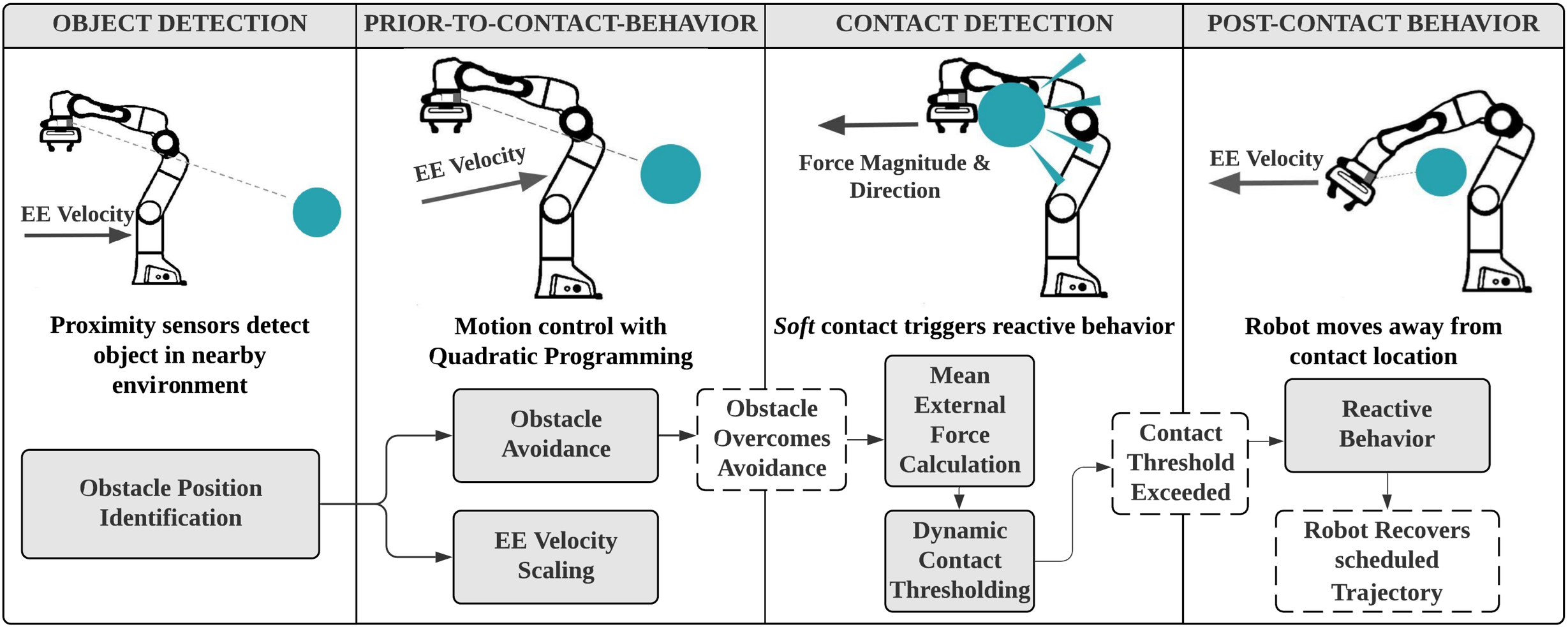}
    \caption{Diagram of the presented framework for contact anticipation, detection, and reaction. Each module represents a distinct interaction from when an object enters the robot's environment to when the repulsive velocity becomes zero, and the initial trajectory is resumed.}\label{fig:framework}
\end{figure*}

In this work, we present a framework to enable safe physical interaction between a human and robot manipulator. 
The system is designed to allow for object avoidance, collision anticipation, and deliberate human contact.
In this section, we detail each component of our framework, which consists of perception, avoidance, contact detection, and post-contact reaction as seen in \cref{fig:framework}.

\subsection{Sensor Units (SUs) equipped with proximity sensing}\label{sec:development_of_artificial_skin_units}
%
%
Real-time physical human--robot interaction can be realized via distributed whole-body sensing. 
To demonstrate this hypothesis, we have developed a self-contained, low-cost, low-power sensor unit (SU) capable of transmitting real-time environmental data wirelessly. The SU (introduced in \cite{watanabe2021calibration}) is designed to be a simple, modular element used to perform prior-to-contact, nearby space perception. 
Each SU utilizes an LSM6DS3 iNEMO 3D accelerometer and 3D gyroscope inertial measurement unit (IMU) for automatic kinematic calibration of the SUs and a VL53L1X time-of-flight (ToF) sensor for distance measurements \cite{watanabe2021calibration}. Distance information is published at a rate of $50$ Hz, with a maximum distance of four meters from the sensor. 
%
The $33$ mm by $36$ mm board in \cref{fig:SU_Contact} is optimized for reduced area and affords the benefits of modularity with a built-in ESP8266 microcontroller with real-time wireless communication capabilities. Our SUs are powered by a $3.7$ V, $1000$ mAh lithium-ion battery and are configurable through onboard programming via a USB-to-UART bridge. Each SU costs approximately \$$36$ and is operational for up to ten hours on a single charge with a current consumption between $110–160$ mA.

\subsection{Identification of Obstacle Positions}\label{sec:identification_of_obstacle_positions}

With our custom sensor units, objects are perceived in the robot's immediate surroundings. To accomplish this, SUs are distributed along the robot's body and automatically located through our previous work's calibration algorithm in \cite{watanabe2021calibration}. 
%
Each sensor unit's proximity sensor is positioned so that its distance measurement is parallel to the sensor unit's $z$-axis (i.e. normal to the robot's surface). Then, the object's position $\mathbf{h_k} \in \mathbb{R}^3$ detected by sensor unit $\mathbf{k}$ as a function of proximity reading $d_{obs, k} \in \mathbb{R}$ can be computed as: 
\begin{equation}
     \mathbf{h_k} = {}^{O}\vec{r}_{SU_k} + {}^{O}{R}_{SU_k}\begin{bmatrix}
0 & 0 & d_{obs, k}
\end{bmatrix}^T ,
\end{equation}
where $SU_k$ is the $k$th sensor unit, ${}^{O}\vec{r}_{SU_k} \in \mathbb{R}^3$ is the position of $SU_k$ with respect to robot base frame $O$, and ${}^{O}{R}_{SU_k} \in \mathbb{R}^{3 \times 3}$ is the rotational matrix of $SU_k$ with respect to the robot base frame.
The $k$th SU's pose information is used to compute an object position in Cartesian space given $d_{obs, k}$. Combined with knowledge of the robot's state, collision avoidance and contact thresholding are now achievable. This section concludes the object detection block of \cref{fig:framework}.

\subsection{Motion Control with Quadratic Programming }\label{sec:motion_control_with_quadratic_programming}

The main task and avoidance behaviors of the system are implemented using a Cartesian velocity controller that solves for joint velocities in a unified quadratic programming (QP) expression. This optimization technique is used to leverage the robot's redundant degrees of freedom to avoid contact while maintaining the desired main task.
%

To define the main task, first, let $\mathbf{\dot{q}} \in \mathbb{R}^n$ represent the joint velocities of a kinematically redundant robot manipulator with $n$ joints, and let the Cartesian velocity of the end-effector (EE) be represented as $\mathbf{\dot{x}} \in \mathbb{R}^m$, where $\mathbf{\dot{x} = J(q) \dot{q}}$, and $\mathbf{J(q)} \in \mathbb{R}^{m \times n}$ is the Jacobian of the robot. With this in mind, the control equation is written as:
\begin{equation} \label{eq:standard_minimization}
\begin{split}
g(\mathbf{\dot{q})} & = \frac{1}{2}\mathbf{(\dot{x}_d - J\dot{q})^{\top}(\dot{x}_d - J\dot{q}) +} \frac{\mu}{2}\mathbf{\dot{q}^{\top}\dot{q}}\\
  &\;\;\;\;  + \frac{k}{2}(\mathbf{\dot{q}_{mid}} - \mathbf{\dot{q}})^{\top}(\mathbf{\dot{q}_{mid}} - \mathbf{\dot{q}}).
\end{split}
\end{equation}
The first term in \cref{eq:standard_minimization} represents the squared Cartesian error between the current EE velocity $\mathbf{\dot{x}}$ and the desired task velocity $\mathbf{\dot{x}_d}$, 
whereas the second term is used as a damping term to avoid singularities based on a robot manipulability measure---we refer the reader to \cite{nakamura1986inverse} for more details. Additionally, a third term is added to the main task that causes the robot to favor joint positions near the middle of its joint limits. In this term, $\mathbf{\dot{q}_{mid}}$ consists of the desired joint velocities that will move the joints towards their middle positions (as determined from the joint operational range from the robot's manufacturer), and $k$ is a scaling factor used to weight the middle joint term. To calculate the $\mathbf{\dot{q}_{mid}}$ value, we compute the difference in the current and middle joint positions, then use that to determine joint velocities that move towards their middle positions over $t$ seconds. 
\cref{eq:standard_minimization} is then manipulated to conform with the quadratic programming notation and restrictions inspired by \cite{ding2020collision}. The previous formulation, which did not include the middle joint limit term, allowed the manipulator to enter undesirable configurations, and ultimately error states.
In the following two sections, we introduce two safety components: end-effector (EE) velocity reduction and robot movement restrictions, which form the constraints of the QP optimization formulation.

\begin{figure*}\vspace{+5pt}
  \centering
  \begin{subfigure}[t]{0.32\textwidth}\centering
    \includegraphics[height=1.6in]{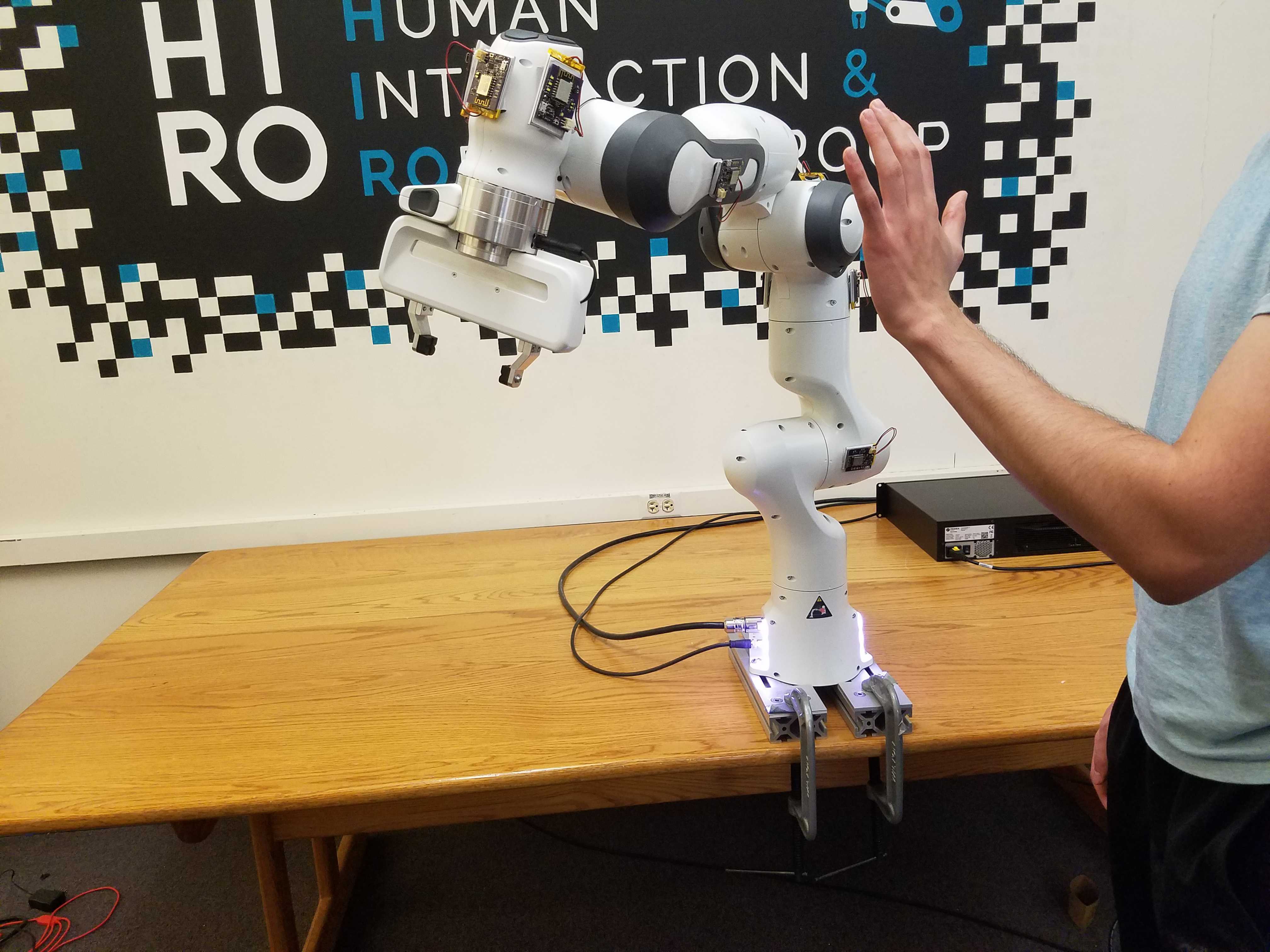}
    \caption{\textbf{Prior-to-Contact:} A human enters the robot's workspace and is sensed by onboard sensor units, triggering both velocity scaling and avoidance behavior.}\label{fig:prior_to_contact}
  \end{subfigure}\hfill
  \begin{subfigure}[t]{0.32\textwidth}\centering
    \includegraphics[height=1.6in]{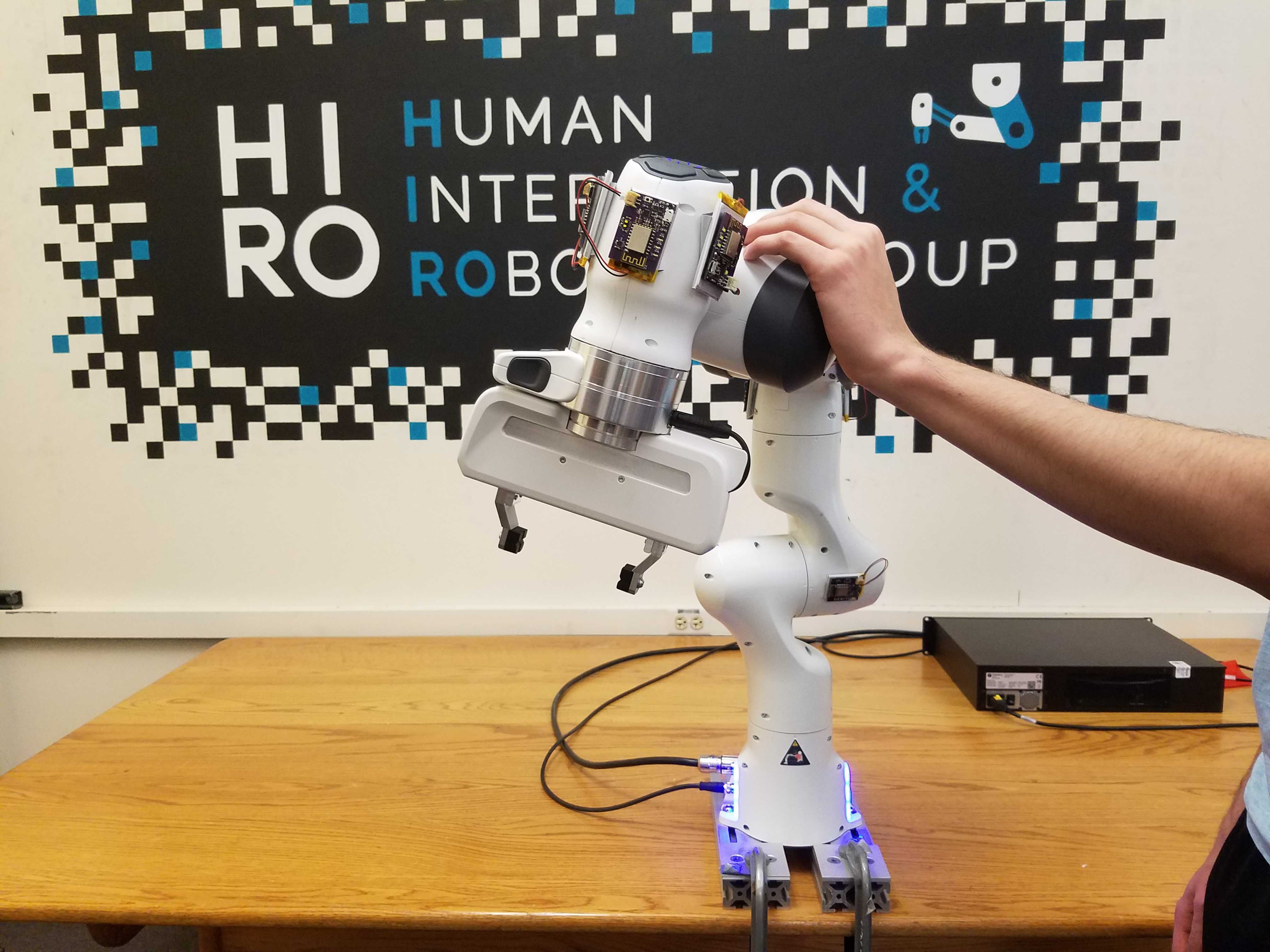}
    \caption{\textbf{Contact:} When an object makes contact with the robot, the external force is measured and used to determine if it exceeds the dynamically computed threshold.}\label{fig:contact}
  \end{subfigure}\hfill
  \begin{subfigure}[t]{0.32\textwidth}\centering
    \includegraphics[height=1.6in]{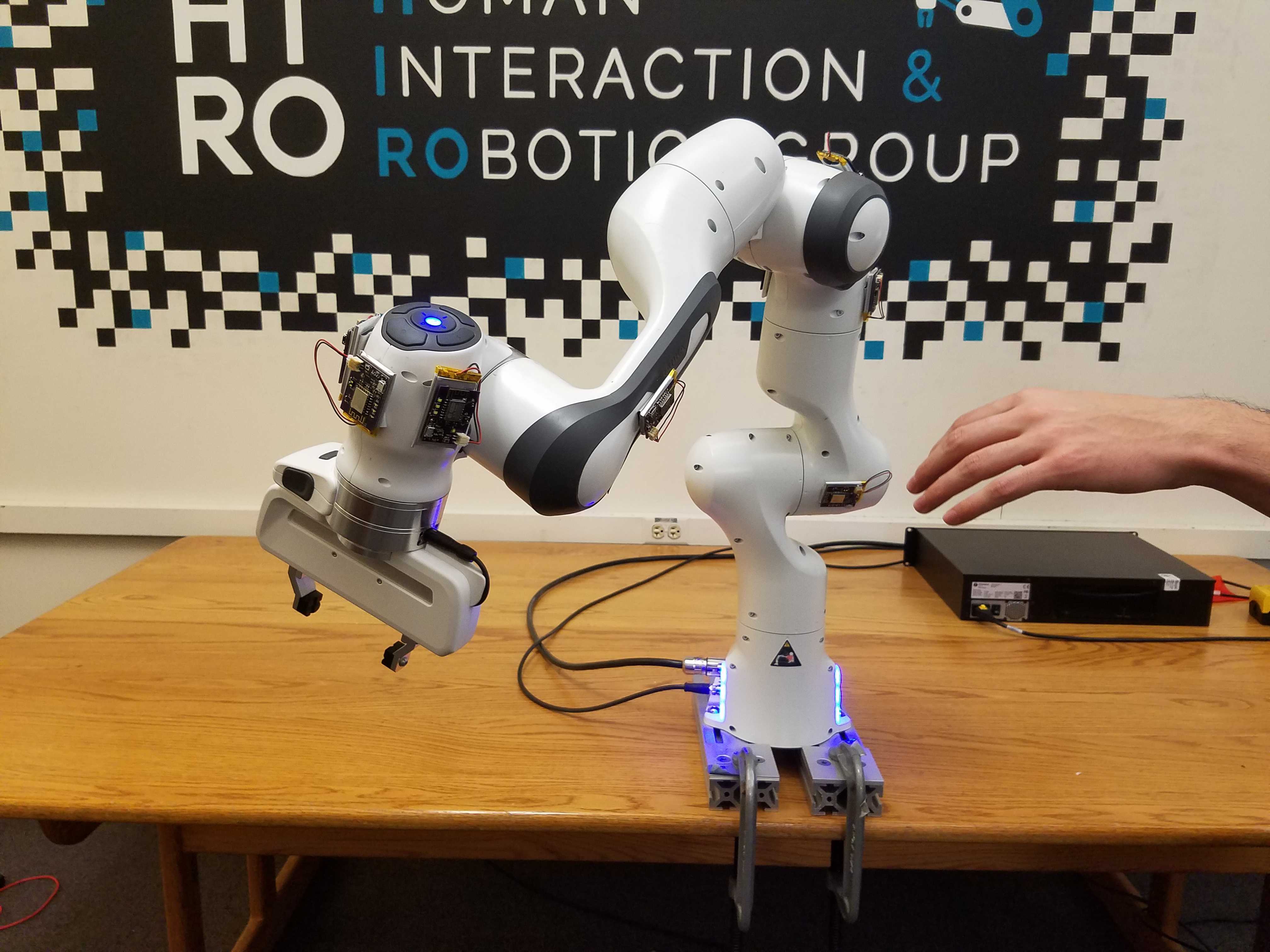}
    \caption{\textbf{Post-Contact:} After the external force threshold exceeded, a reactive behavior moves the robot in the direction and magnitude of the applied force.}\label{fig:post_contact}
  \end{subfigure}
  \caption{A human physically interacting with a robot while in motion. This interaction is outlined in \cref{sec:dynamic_obstacle_collision}.}\label{fig:human_robot_experiment}
\end{figure*}
\vspace{-3pt}
\subsection{End-Effector Velocity Reduction}\label{sec:end_effector_velocity_scaling}
When a robot detects an obstacle, its EE velocity is reduced in order to maximize safety and prepare for contact. As a human approaches, the robot slows down execution of its main task, which causes the corresponding avoidance motion to be slower; this allows contact to easily be made. Without any EE velocity reduction, purposeful contact is both difficult and unnatural.
%
The EE velocity is reduced as follows: for each detected obstacle $\mathbf{h_k}$ within a user-defined max distance $d_{max}$, the norm of the distance to the robot EE is taken, $||d||$. We then take the smallest distance norm, $||d_{lowest}||$, to calculate the reduced EE velocity shown in the following equation:
%
%
%
\begin{equation}
    \mathbf{\dot{x}_d} = \frac{||d_{lowest}||}{d_{max}}\ \mathbf{\dot{x}_d}.
    \label{eq:xd}
\end{equation}
Above, $\frac{||d_{lowest}||}{d_{max}}$ produces a scaling term that reduces the main task's desired EE velocity, dependent on the distance to the closest object. Objects detected beyond $d_{max}$ from the EE are discarded. Furthermore, to prevent jerky motions induced by vanishing obstacle readings, we apply a linear decay formula to simulate an obstacle slowly moving away from the robot:
%
%
%
\begin{align}
    \mathbf{\dot{x}_d} = \mathbf{\dot{x}_d} \cdot \left[\frac{||d_{lowest}||}{d_{max}} + (1 - \frac{||d_{lowest}||}{d_{max}}) \cdot \frac{l_{obs}}{l_{max}}\right],
    \label{eq:linear_reduction}
\end{align}
where $l_{obs}$ is a term which starts at 0 and linearly increases to $l_{max}$, and $l_{max}$ is the value of where the original EE velocity is completely restored. All user-defined values are located in \cref{table:parameters}.
\vspace{-1pt}
\subsection{Robot Movement Restrictions}\label{sec:movement_restrictions}

Movement restrictions are applied by adding optimization constraints to the QP main task formulation. With these constraints, a robot moves \textsl{around} objects instead of directly \textsl{away} from them. This is distinctly different than EE velocity reduction; here, we restrict the velocity of dynamically determined control points along the robot's body closest to each obstacle.
A linear inequality constraint is constructed that limits the motion $\dot{\mathbf{x}}_{\mathbf{c, i}} \in \mathbb{R}^3$, which is the Cartesian velocity of the closest point to obstacle $\mathbf{i}$ on the manipulator. This value is computed as $\mathbf{\dot{{x}}_{{c, i}} = J_{c, i}{\dot{q}}}$, where $\mathbf{J_{c, i}}$ is the Jacobian of the closest point on the manipulator to the obstacle. 
Next, $\mathbf{\hat{d}}$ is computed, which describes the direction from the dynamic control point to the obstacle $\mathbf{h_i} \in \mathbb{R}^3$; thus, the term $\mathbf{\hat{d}^T J_{c, i}{\dot{q}}}$ describes the approach velocity of the closest point on the robot towards the object. 
Finally, the control point's approach velocity towards the closest object is constrained by a maximum approach velocity, $\dot{x}_a$, which is determined based on the distance to the closest obstacle $\mathbf{h_i}$.
We compute $\dot{x}_a$ as a smooth and continuous function.
%
Drawing inspiration from \cite{flacco2012depth}, our movement restrictions are computed as follows:
\begin{equation}
    V_a = \frac{V_{max}}{1 + e^{\beta(2\frac{d}{d_{crit}} - 1)}};
    \label{eq:Va}
\end{equation}
\begin{equation}
    V_b =  \frac{V_{max}}{1 + e^{\beta(2\frac{d - d_{crit}}{d_{notice} - d_{crit}} - 1)}};
    \label{eq:Vb}
\end{equation}
\begin{gather}
        \dot{x}_a = \begin{cases}
         V_a- V_{max} : \text{if } d < d_{notice} \text{ and } d < d_{repulse}, \\
       V_b: \text{if } d < d_{notice} \text{ and } d \geq d_{repulse}, \\
        \text{Drop Restriction}: \text{otherwise}.
        \end{cases}
        \label{eq:approach_velocity_computation} 
\end{gather}
$V_{max}$ is the maximum repulsive velocity, $d$ is the distance from the object to the closest point on the robot body, $d_{repulse}$ is the distance where repulsive behavior begins. $d_{notice}$ is the distance at which we ``notice'' an obstacle and start to impose movement restrictions. 
When an object is close, a small repulsive velocity is applied --- the robot will only slowly avoid the object, and contact can be easily made if desired. The parameters in the previous three equations can be found in \cref{table:parameters}.
%
%
\vspace{-3pt}
\subsection{Dynamic Contact Thresholding}\label{sec:dynamic_contact_thresholds}
\vspace{-2pt}
In order to smoothly transition between prior-to-contact and post-contact behaviors, a controller must determine if external contact has been made, along with the direction and magnitude of that contact. Three qualities are desired to perform a seamless transition into post-contact behavior: a) perfect external force data should not be required, b) contact is more likely when an object is close to the robot, and c) a robot moving close to an object should move slower than normal and have \textbf{increased} sensitivity to interaction. To this end, we propose a dynamic contact thresholding algorithm that relies on obstacle readings and an estimation of the robot's external force to guarantee all three desired qualities.

To detect contact forces, the proposed algorithm does not require a perfect estimation of the external force: the signal that we utilize throughout this paper is the estimated external Cartesian contact force, provided through the Franka Control Interface \cite{frankacontrolinterface}. This estimation is often volatile and far from zero in all axes, especially when the robot is moving at high velocities. However, our algorithm is robust to noisy data and can be used to accurately determine if external contact is made on the robot's EE.

\subsubsection{Average External Force Calculation}

Given estimated external force data, a sliding window is used to determine the running average.
Values are added to the window as detailed below:
\begin{gather}
        x_t = \begin{cases}
        \alpha x_t  + (1 - \alpha)x_{t-1}: \text{if } |x_t - \mu_{t-1}| > \lambda \sigma_{t-1},  \\
        x_t: \text{otherwise},
        \end{cases}
        \label{eq:weighted_new_value}
    \end{gather}
where $x_t$ is the data point appended to the window, $\alpha$ is the influence value of a detected outlier, $\lambda$ is a scaling factor that is multiplied by the previous standard deviation $\sigma_{t-1}$, and $\mu_{t-1}$ is the previous mean value. The term $\lambda \sigma_{t-1}$ is used to determine if a given value is an outlier and should be discounted when added to the moving average. 
%
%
\subsubsection{Contact Threshold Calculation}
Contact thresholding determines when contact is made, in what directions contact was made, and the magnitude of the contact. To start, from the external force window, we use the running average ($\mu$) as a base value for the dynamic external contact thresholds. For each axis, we compute an independent contact threshold, which is expressed as:
\begin{align}
    \text{Contact Threshold} =
    \begin{cases}
    T_u = \mu + F_b + F_{\sigma} - F\SPSB{--}{obs}, \\
    T_l = \mu - F_b - F_{\sigma} + F\SPSB{+}{obs},
    \end{cases}
    \label{eq:contact_threshold}
\end{align}
where $T_u$ is the upper and $T_l$ is the lower limit of the contact threshold, and $F_b$ is the base additional force required from the mean to trigger contact behavior. $F_{\sigma}$ increases the external force necessary to trigger contact behavior and is based on the external force signal's standard deviation. This value is computed as 
\begin{gather}
    F_{\sigma} = \min(\frac{\sigma}{\sigma_{max}} \cdot F_{std}, F_{std}), 
    \label{eq:F_sigma}
\end{gather}
where $\sigma_{max}$ is the max standard deviation. $\sigma$ is the external force window's current standard deviation, and $F_{std}$ is the max amount of force that is applied based on the $\sigma$ value. From the above equation, when the standard deviation of the current window is low, $F_{\sigma}$ will be small, which leads to a threshold that is sensitive to external forces. As the robot slows down, the external force data is less noisy, making a slower robot more sensitive to contact. Next, $F\SPSB{-}{obs}$ and $F\SPSB{+}{obs}$ are contact force reductions of the upper and lower limits, respectively, based on the distance from an obstacle to the robot's EE. An object close to the robot EE reduces the contact threshold because contact is likely when objects are nearby.
$F\SPSB{-}{obs}$ and $F\SPSB{+}{obs}$ are computed using the following equation:
\begin{gather}
      F_{\text{obs}} = (\dfrac{d_{\text{max}} - d}{d_{\text{max}} - d_{\text{min}}}) \cdot F_{d},
      \label{eq:F_obs}
\end{gather}
where $d_{max}$ is the distance from the robot's EE where $F_{\text{obs}}$ begins to be applied, and $d_{min}$ is the distance at which the maximum force reduction is applied. $F_d$ is the maximum force reduction, and $d$ is the minimum distance from the robot's EE to an object on the positive side of the robot for $F\SPSB{+}{obs}$ and the negative for $F\SPSB{-}{obs}$. For example, an object with a greater $y$ value than the current end-effector position will lead to a reduction in the required force to trigger contact behavior in the negative $y$ direction.

If the newest value added to the running average exceeds either the upper or lower threshold in any one or more of the robot's axes, then, a reactive behavior will occur.
We calculate the direction and overall external force applied to the EE and use that information to apply a velocity in the external force's direction, as shown in
\begin{align}
      F_{ext} = F_{reading} - \mu, \qquad \dot{\bf{x}}_{des} = C \cdot F_{ext}.
      \label{eq:F_ext}
\end{align}
$F_{reading} \in \mathbb{R}^3$ describes the contact force from the sliding window, and $\mu \in \mathbb{R}^3$ is the mean force of the window, which computes $F_{ext} \in \mathbb{R}^3$. $C \in \mathbb{R}^{3 \times 3}$ describes the Cartesian compliance matrix that proportionally multiplies each component of the external force to output a desired end-effector velocity, which temporarily overrides the main task. 
We then linearly decay the velocity to zero over a specified time period, similar to \cref{eq:linear_reduction}.\\  

Ultimately, as a result of dynamic contact thresholding, the three aforementioned interaction qualities are achieved:\vspace{+2pt}
\begin{enumerate}
    \item \textbf{Robustness to measurement noise: } The controller does not rely on perfect external force data; instead, it is able to detect contact from noisy, external forces.\vspace{+2pt}
    \item \textbf{Anticipated contact from obstacles: } As an obstacle approaches the robot's EE, the contact thresholds decrease due to the $F_{\text{obs}}$ value.\vspace{+2pt}
    \item \textbf{Increased contact sensitivity as velocity decreases:} For noisier data, we are less certain of contact being made, which expands the threshold bounds. When a robot slows down due to obstacles, reduced noise is observed, and, in turn, a smaller $F_{\sigma}$ value is added to the thresholds.\vspace{+2pt}
\end{enumerate}

\section{Experiment Design}\label{sec:experiment_design}
\begin{table}\vspace{+5pt}
\centering
\begin{tabular}{|c|c|c|} 
 \hline
 Parameter & Value  & Equations\\ [0.5ex] 
 \hline\hline
%
 $d_{max}$ & 0.8 m  & \ref{eq:xd}, \ref{eq:linear_reduction}, and \ref{eq:F_obs} \\ 
 \hline
 $d_{min}$ & 0.05 m  & \ref{eq:F_obs} \\
 \hline
 $l_{max}$ & 200 cycles @ 100Hz & \ref{eq:linear_reduction}\\
 \hline
 $V_{max}$ & 0.04 m/s & \ref{eq:Va}, \ref{eq:Vb}, and \ref{eq:approach_velocity_computation}\\
 \hline
 $d_{repulse}$ & 0.1 m  & \ref{eq:approach_velocity_computation}\\
 \hline
 $d_{crit}$ & 0.1 m  & \ref{eq:Va} and \ref{eq:Vb}\\ 
 \hline
 $d_{notice}$ & 0.6 m  & \ref{eq:Vb} and \ref{eq:approach_velocity_computation}\\
 \hline
 $\alpha$ & 0.1  & \ref{eq:weighted_new_value}\\ 
 \hline
  $\beta$ & -10  & \ref{eq:Va} and \ref{eq:Vb} \\ 
 \hline
 $\lambda$ & 0.75  & \ref{eq:weighted_new_value}\\ 
 \hline
 $F_{std}$ & 3 N & \ref{eq:F_sigma}\\
 \hline
 $F_d$ & 4  N & \ref{eq:F_obs}\\
 \hline
 $F_b$ & 10 N & \ref{eq:contact_threshold}\\
 \hline
 $\sigma_{max}$ & 3 & \ref{eq:F_sigma}\\
 \hline
\end{tabular}
\caption{Table of user-defined parameters used in the evaluation of the proposed method.}
\label{table:parameters}
\end{table}
To validate the effectiveness of our algorithm, experiments are conducted on a physical, 7-DoF Franka Emika Panda robotic arm. A computer outfitted with a real time kernel  updates the commanded robot joint velocities at a rate of $100$ Hz. The control loop acts on the most recently available sensor unit data. We utilize a kinematic calibration algorithm \cite{watanabe2021calibration} and sensor units detailed in \cref{sec:development_of_artificial_skin_units} to detect objects. Each sensor unit is outfitted with an IMU and proximity sensor, the latter of which informs our avoidance and contact detection algorithms.
The experimental pipeline is developed using C++ and ROS \cite{quigley2009ros}: we initially implemented our avoidance algorithms in the Gazebo simulator \cite{koenig2004design}, then transitioned to the real robot to test the entire framework. We evaluate our framework in multiple scenarios, as detailed in the next sections. Lastly, a list of explicit user-defined parameters can be found in \cref{table:parameters}.
\subsection{Static Obstacle Collision} \label{sec:static_obstacle_collision}
In the first scenario, the robot is placed near a large static object and commanded to move towards it, causing a collision. Proximity information is used to slow down the robot and enable soft contact. Specifically, we position the object in two positions around the robot to highlight the ability of our system to reduce contact thresholds based on object location. In the first position, the static object is placed in the negative $y$ direction of the end-effector; in the second position, it is placed in the positive $y$ direction.
Two different trajectories are used for each collision -- in the first, the robot moves in a Cartesian circle with a radius of $0.25$ m, centered at ($0.5$, $0, 0.25$), and moving counter-clockwise in the $y$ and $z$ direction. For the second, the robot moves in a line from ($0.4$, $0.5$, $0.5$) to ($0.4$, $-0.5$, $0.5$). Both trajectories are commanded to move at a max speed of $0.3$ m/s when no objects are nearby. Each of these trajectories are compared to that of the same movement when not informed by onboard SUs to anticipate contact and reduce movement speed.
\begin{figure}\vspace{+5pt}
    \centering
    \includegraphics[width=0.84\linewidth, height=6cm]{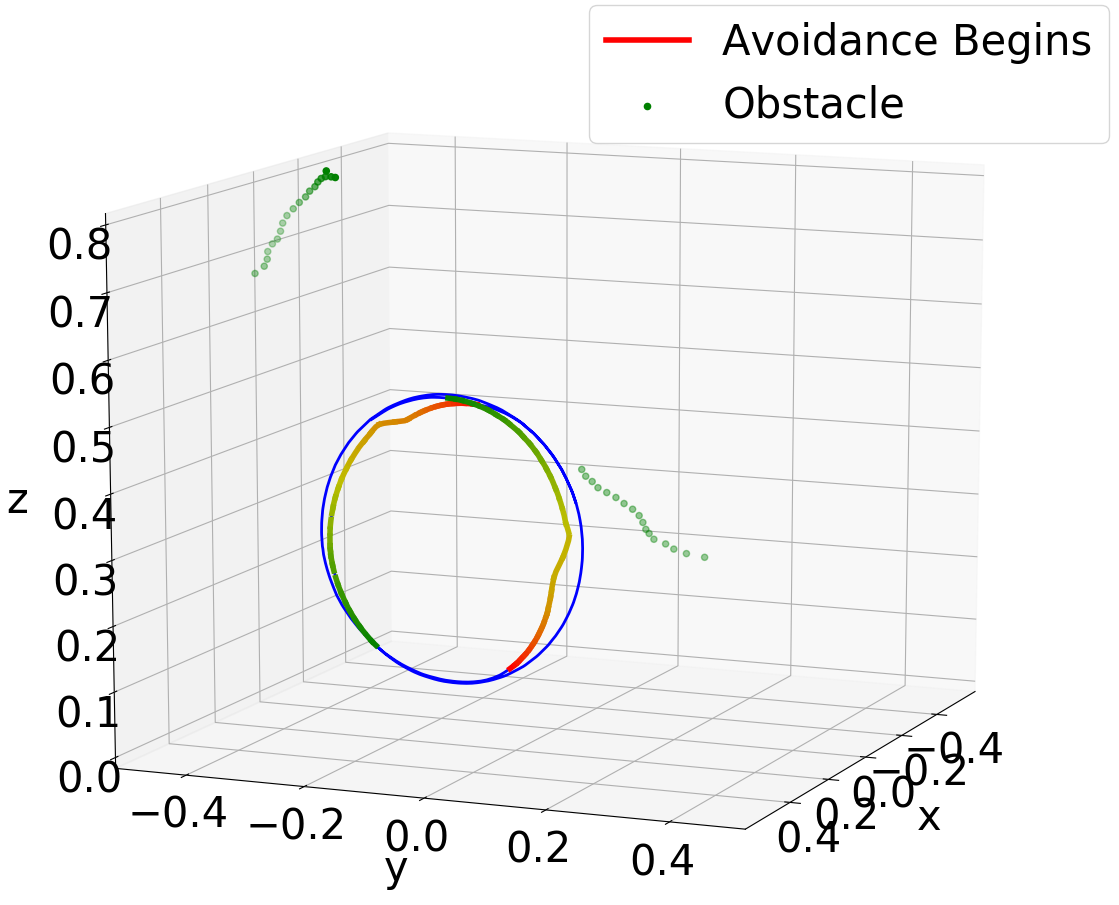}
        \caption{Robot EE moving in a circular path (blue circle); when an object is detected (green dots), the end-effector trajectory is modified to avoid collision (red) then gradually recovers (green). We demonstrate this behavior in two interactions on opposite sides of the robot; the objects are detected by independent proximity sensors.}
    \label{fig:ee_trajectory}
\end{figure}
\begin{figure*}\vspace{+5pt}
    \centering
    \includegraphics[width=0.92\linewidth]{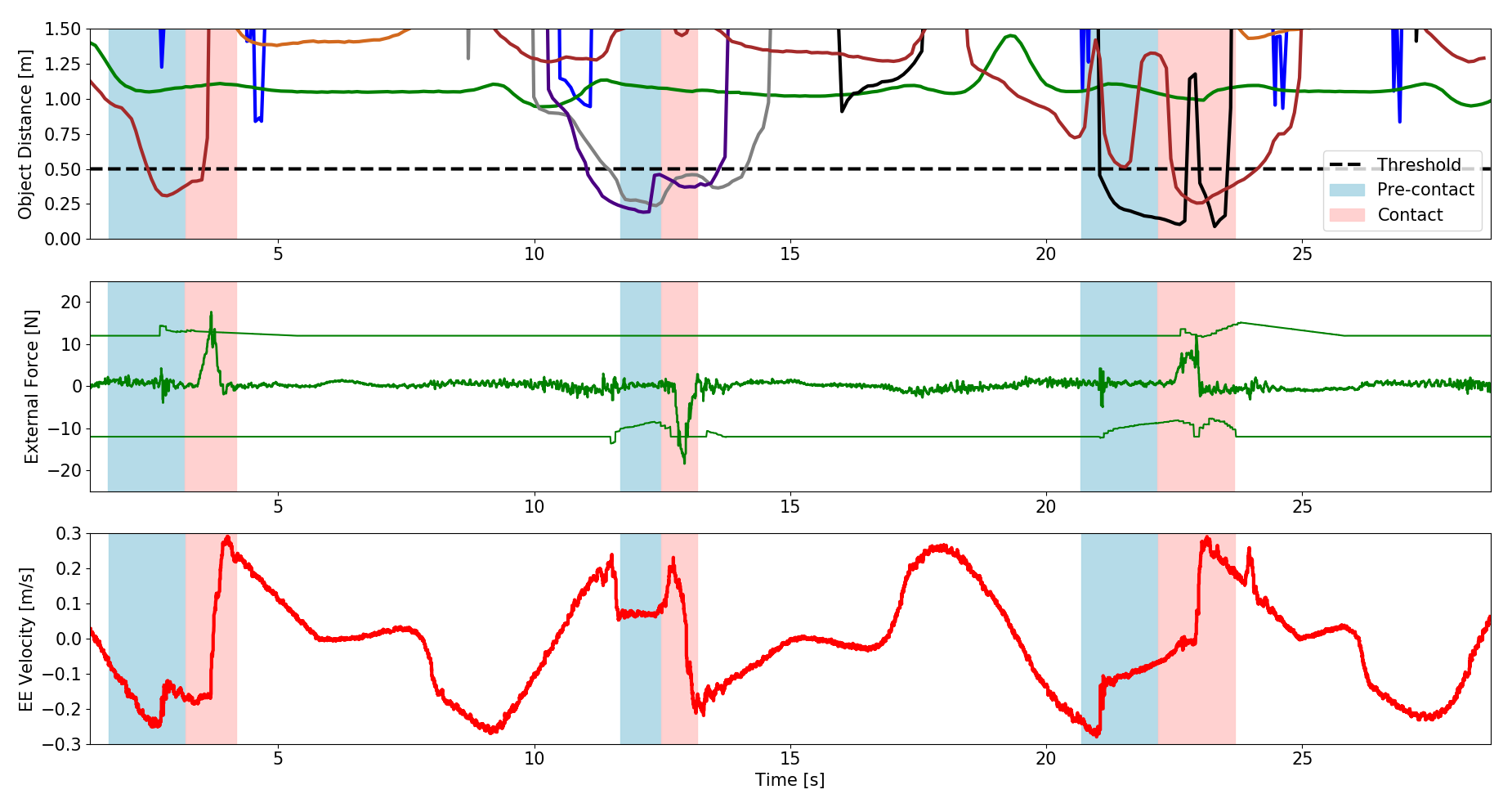}
        \caption{Three components of a continuous human--robot interaction over $30$ seconds: during this time, three distinct contact are made from multiple directions. Top graph: depth values of seven proximity sensors placed along the robot body and EE, with the activation threshold as a dotted black line. Middle graph: external force exerted on the robot EE in the y-axis with the dynamic thresholds shown above and below. Bottom graph: The manipulator's EE velocity along the $y$-axis.}
    \label{fig:dynamic_obstacle_collision}
\end{figure*}
\subsection{Obstacle Avoidance} \label{sec:obstacle_avoidance}
In this scenario, the robot is commanded to move in the same circular trajectory as in \cref{sec:static_obstacle_collision}, while a human enters the robot's path. When the human enters the scene, they cause a slight deviation in the robot's initial circular trajectory. An example of this is illustrated in \cref{fig:prior_to_contact}, where a human holds their hand up near the robot's end-effector.
\subsection{Dynamic Obstacle Collision} \label{sec:dynamic_obstacle_collision}
Finally, multiple dynamic obstacle collisions with a human participant are made to demonstrate the full capabilities of our system. Through this interaction, multiple sensors detect objects and contact is made from different directions. After contact has been made and the reactive behavior has been completed, the robot continues along its original trajectory. A depiction of this interaction can be seen in \cref{fig:human_robot_experiment}.

\section{Results and Discussion}\label{sec:results}
\subsection{Static Obstacle Collision}
\label{sec:static_obs_results}
\begin{figure}
    \centering
    \begin{subfigure}{0.45\textwidth}
        \centering
        \includegraphics[width=1.0\textwidth]{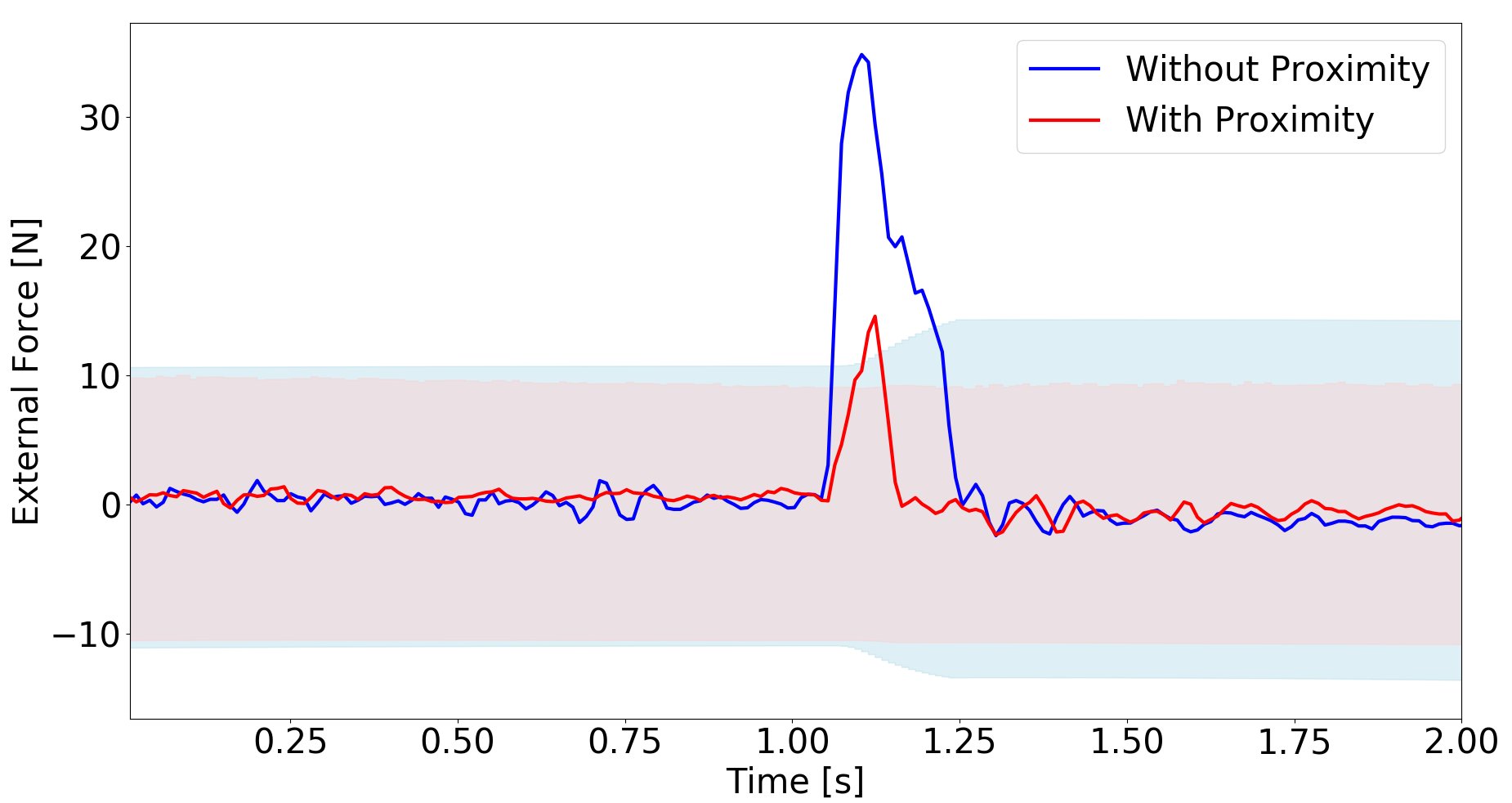}
    \end{subfigure}
    \begin{subfigure}{0.45\textwidth}
        \centering
        \includegraphics[width=1.0\textwidth]{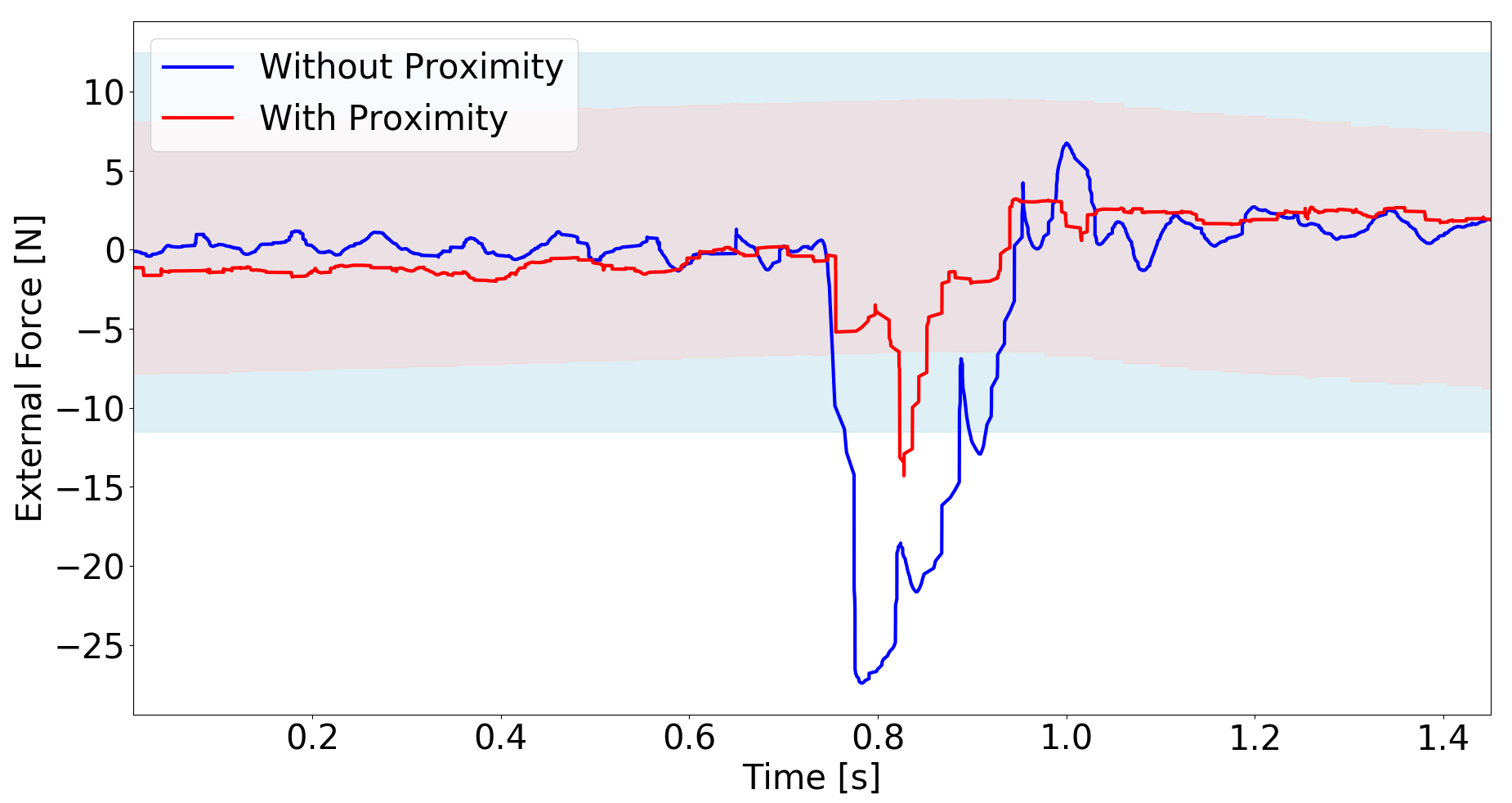}
    \end{subfigure}
    \caption{Force difference for contact detection, in the $y$-direction, between our method with proximity sensing (red line) and without proximity sensing (blue line). The shaded regions represent the thresholds for contact. In the top graph, the robot is moving in a circular trajectory, and in the bottom, a horizontal line, before contact is made.}
    \label{fig:contact_differences}\vspace{-12pt}
\end{figure}

\cref{fig:contact_differences} illustrates the force reduction when proximity sensors are used to anticipate a collision. In both the circular and linear trajectories, the difference in the overall detected contact force is approximately two times smaller when utilizing proximity sensor data. The smaller forces demonstrate that proximity detection for collision anticipation can make close proximity interactions safer.

Prior to contact, the contact threshold value in the object's direction is reduced, as seen most prominently in the concave red shaded region of the bottom graph in \cref{fig:contact_differences}. This change in threshold is due to object proximity and EE velocity scaling, which are effectively able to soften the collision force between the robot and object. Two independent contacts in both the robot's positive and negative $y$-axis are displayed in \cref{fig:contact_differences}. While we purposefully allow our robot to make contact with a static object, the robot's avoidance weights can easily be altered to impose harsher movement restrictions that completely avoid the object. Dependent on the task, these parameters can be tuned to fit a certain use case.
\begin{figure}[h!]
    \centering
    \includegraphics[width=1.0\linewidth]{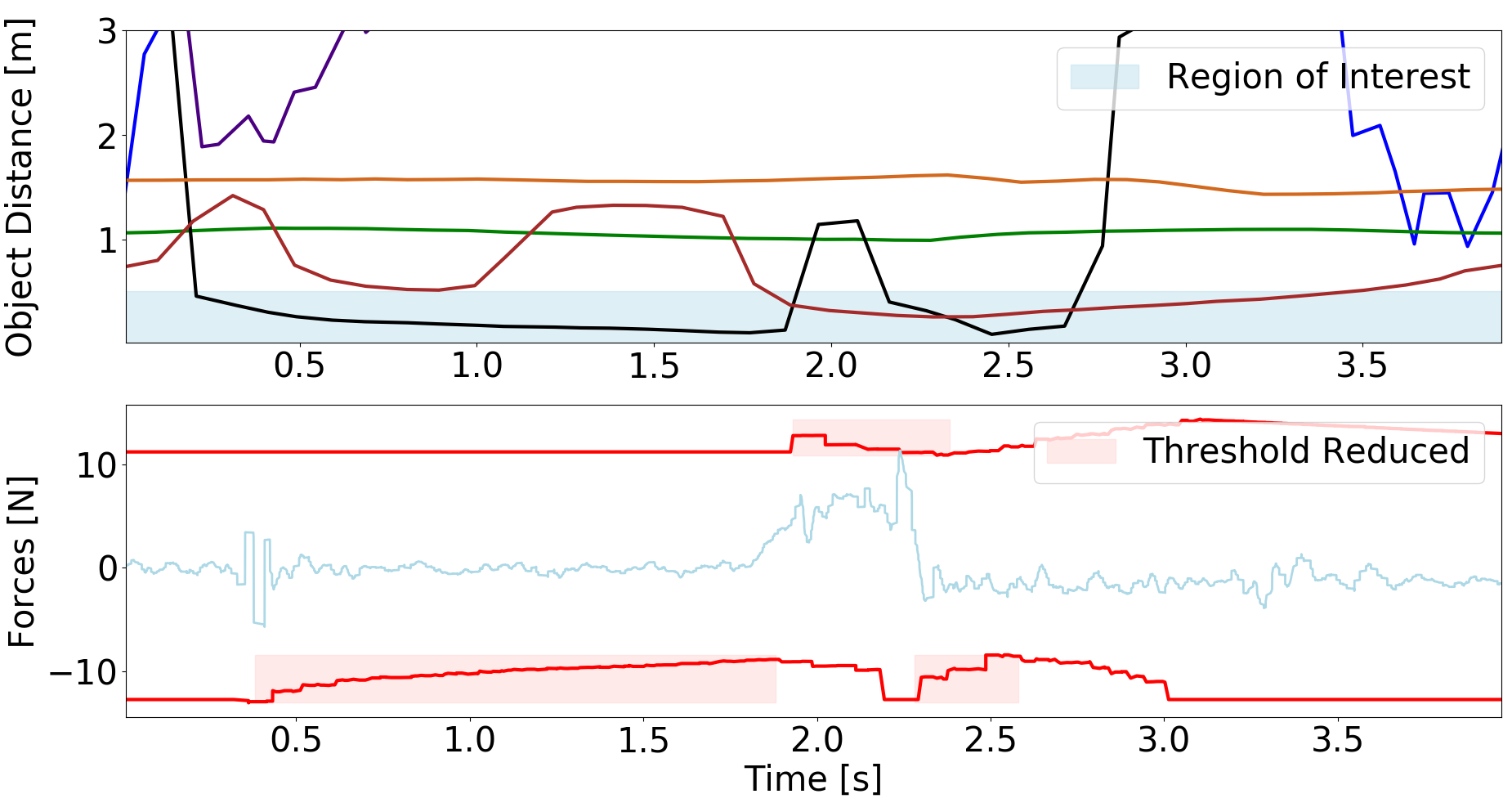}
        \caption{Top graph: distance values from six proximity sensors mounted on the robot body and EE during the interaction outlined in \cref{sec:dynamic_obstacle_collision}. The region of interest label denotes the region at which sensed distances are meaningful for our algorithm. Bottom graph: external force exerted on the robot EE in the $y$ direction and associated contact thresholds above and below.}
    \label{fig:contact_threshold}\vspace{-16pt}
\end{figure}
\subsection{Obstacle Avoidance}
\cref{fig:ee_trajectory} shows the robot's circular trajectory altered when an object enters its nearby space environment and triggers an avoidance. As an object appears in the robot's operational space, the controller scales the its velocity, and causes a slight deviation in the circular path through movement restrictions. Once far enough away from, or moving in the opposite direction of the obstacle, the robot returns to its original trajectory. We showcase this behavior using multiple SUs mounted on opposite sides of the manipulator to emphasize that the controller can simultaneously avoid collisions and anticipate contact. Similar to \cref{sec:static_obs_results}, obstacle avoidance parameters can be tuned to increase or decrease restrictive behavior.
\subsection{Dynamic Obstacle Collision}
The final evaluation highlights multiple dynamic contacts with a human as outlined in \cref{fig:human_robot_experiment}. These interactions are detailed in \cref{fig:dynamic_obstacle_collision}, three separate contacts are made with the manipulator. The robot's movement is only affected by obstacles with distance less than a predetermined threshold of $0.5$ m (the diameter of its circular trajectory), seen as a black dotted line. When the distance condition is met, the robot's velocity is reduced, as seen in the blue shaded region of the bottom graph in \cref{fig:dynamic_obstacle_collision}. In addition to the robot slowing down, the contact thresholds become more restrictive when objects are nearby; this is most clearly shown in \cref{fig:contact_threshold}, which is a higher resolution view of the third interaction of \cref{fig:dynamic_obstacle_collision}. During the interaction in \cref{fig:contact_threshold}, two proximity sensors are activated on opposite sides of the manipulator, and this leads to a reduction of both upper and lower thresholds, as seen in the red shaded region. After slowing down, contact with the robot is shown in the red shaded regions of \cref{fig:dynamic_obstacle_collision} from multiple directions. The direction and magnitude of the applied external force directly determines the direction and magnitude of the EE velocity when repulsive behavior is triggered. We can see the resulting velocity spikes from contact in the bottom graph of \cref{fig:dynamic_obstacle_collision} in the red shaded regions. Through this interaction, we illustrate that our system is able to anticipate and react to contact with dynamic objects within its environment from multiple directions. In fact, without our thresholding behavior, such soft contact would not be discernible and not allow a human to easily interact in close proximity. The implicit anticipation opens a new realm of behaviors for true human--robot collaboration in close proximity, where a robot can dynamically adapt to its changing environment.
\section{Conclusions}\label{sec:conclusions}
In this work, we introduce a framework for contact anticipation during physical human--robot interaction. 
Through relaxed avoidance constraints in a quadratic programming control formulation, combined with a novel dynamic thresholding algorithm, our work addresses an unexplored area between avoidance and contact.
Our experiments demonstrate the system's abilities in multiple scenarios with both static and dynamic interactions.

With respect to future work, we plan to develop a truly collaborative system through both hardware and software. Our current sensor units are outfitted with IMUs; while we used the IMUs for our calibration, they were not utilized in this framework. IMU data can be leveraged to increase sensitivity to external forces acting on the robot. The next iteration of sensor units will include tactile sensors, allowing for \textsl{localization} of contact from external forces. Another promising direction for future research is a parametric analysis of the proposed algorithm, where we can explore the sensitivity of our framework's parameters.
Additionally, we plan to explore the effectiveness of our perception as we add more sensor units along a robot's exterior and combine this information with external vision to gain a holistic view of the robot's environment. As we move in these future directions, we also plan to explore the differences between simulation and real life to mitigate the challenges of deployment on real robots.
Ultimately, we will continue to improve our current sensor units and achieve whole--body awareness, effectively bringing the paradigms of avoidance and contact together for physical human--robot interaction.

\addtolength{\textheight}{-4cm}   


\bibliographystyle{IEEEtran}
\bibliography{root}

\end{document}